\documentclass[letterpaper, 10 pt, conference]{ieeeconf}  % Comment
\usepackage[noadjust]{cite}
\usepackage{filecontents}
\usepackage{cite}

\usepackage{epsfig}
\usepackage{graphicx}
\usepackage{amsmath}
\usepackage{amssymb}
\usepackage{mathrsfs}
\usepackage{url}

\usepackage{algorithm}
\usepackage{float}
\usepackage{subfiles}
\usepackage{nth}
\usepackage[noend]{algpseudocode}

\IEEEoverridecommandlockouts                              % This command is only needed if 
\title{\LARGE \bf Semi-Semantic Line-Cluster Assisted Monocular SLAM for Indoor Environments
}

\author{Ting Sun$^{1}$, Dezhen Song$^{2}$, Dit-Yan Yeung$^{3}$, Ming Liu$^{1}$% <-this % stops a space
\thanks{$^{1}$ Department of Electronic \& Computer Engineering, Hong Kong University of Science and Technology, Hong Kong, China. {\tt\small (tsun, eelium)@ust.hk} }
\thanks{$^{2}$ Department of Computer Science and Engineering, Texas A\&M University, Texas, US. {\tt\small dzsong@cse.tamu.edu}}
\thanks{$^{3}$ Department of Computer Science, Hong Kong University of Science and Technology, Hong Kong, China. {\tt\small dyyeung@cse.ust.hk}}%
}

\begin{document}

\maketitle
\thispagestyle{empty}
\pagestyle{empty}

%%%%%%%%%%%%%%%%%%%%%%%%%%%%%%%%%%%%%%%%%%%%%%%%%%%%%%%%%%%%%%%%%%%%%%%%%%%%%%%%
\begin{abstract}
This paper presents a novel method to reduce the scale drift for indoor monocular simultaneous localization and mapping (SLAM).  We leverage the prior knowledge that in the indoor environment, the line segments form tight clusters, e.g. many door frames in a straight corridor are of the same shape, size and orientation, so the same edges of these door frames form a tight line segment cluster.  We implement our method in the popular ORB-SLAM2, which also serves as our baseline.  In the front end we detect the line segments in each frame and incrementally cluster them in the 3D space.  In the back end, we optimize the map imposing the constraint that the line segments of the same cluster should be the same.  Experimental results show that our proposed method successfully reduces the scale drift for indoor monocular SLAM.     
\end{abstract}

%%%%%%%%%%%%%%%%%%%%%%%%%%%%%%%%%%%%%%%%%%%%%%%%%%%%%%%%%%%%%%%%%%%%%%%%%%%%%%%%
\section{INTRODUCTION}
\label{sec:introduction}
Visual simultaneous localization and mapping (vSLAM) is widely adopted by robots to explore indoor environments where a GPS signal is unavailable.  There are three dominant types of vSLAM, which are classified according to the sensors used: monocular, stereo and RGB-D.  Compared with stereo and RGB-D, monocular vSLAM is the most hardware-economical and is free of the trouble of calibration and synchronization.  However, monocular vSLAM suffers from scale drift \cite{strasdat2010scale}.  Without any extra sensor, currently this drift is corrected by loop closure with each camera pose represented by 7 DoF (degree of freedom) rather than 6 DoF in the graph optimization \cite{strasdat2010scale,engel2014lsd}.  There are three major limitations of purely relying on loop closure: 1) the robot may never return to the same place; 2) the same place may have a different appearance due to different lighting conditions, viewing angle, occlusion caused by moving objects etc., and 3) some places simply cannot be distinguished merely by appearance, e.g. all the segments of a long corridor could have an identical appearance. 

In this paper we present a novel method to reduce the scale drift for indoor monocular SLAM by leveraging two key observations about the indoor environment and the property of scale drift:

\begin{enumerate}
	\item Buildings are carefully crafted by humans and are full of standardized elements, e.g. all the door frames in a straight corridor are of the same shape, size and orientation; all the bricks tessellating the floor are also the same; etc.  Most of these elements appear very frequently and their edges form tight line segment clusters.  
	\item Scale drift happens gradually, i.e. within the distance of adjacent doors, a robot can still correctly identify the edges of the two doors are of the same length, and this gives the robot a chance to correct the small drift before it accumulates.
\end{enumerate}

Our method identifies and builds line segment clusters incrementally in the front end,  and optimizes the map in the back end by imposing the constraints that the segments of the same cluster should be the same.  We implement our method in ORB-SLAM2 \cite{mur2015orb,mur2017orb}, and the experimental results show that our proposed method successfully reduces the scale drift for indoor monocular SLAM and the whole system runs in real time.  We describe our method as `semi-semantic' since it does NOT fall into the type of using line feature for matching in odometry as many others did, but leverages the observation of `how the line segments appear indoors' and impose global constraints in a SLAM system,  yet this prior knowledge is not as `high level' and `semantic' as commonly mentioned `reappear object' and `scene understanding'.

There are 4 appealing properties of our proposed method:

\begin{enumerate}
	\item Without requiring extra sensors, the scale regulation is explored from frequent appearing standardized indoor elements, i.e. line segments of the same length and orientation.
	\item It does not need pre-training nor access to a model database during running.
	\item It tightly integrates to the ORB-SLAM2 \cite{mur2015orb,mur2017orb} system to achieve the most benefit with the least computation, i.e. it takes advantage of the key points detected by ORB-SLAM2 \cite{mur2015orb,mur2017orb} to filter the line segments' end points, then directly uses the 3D map points to calculate the 3D representation of the line segments.
	\item Compared with approaches using object recognition etc., our method makes a relatively weaker assumption and thus is very generalizable in indoor environments.
\end{enumerate}

Notice that our method is not a replacement of loop closure for the following reasons: 1) they are designed for different application scenarios, i.e. loop closure is for when the robot revisits the same place, while our method is for indoor environments;  2) though both impose regulation in the optimization graph, loop closure only takes effect during revisiting, while our method keeps regulating the SLAM system on its way; 3) their implementations have no conflict nor dependency.  Loop closure is a standard and mature module in SLAM system, it can eliminate the drift once successful.  Our method is a novel approach that leverages the regularity of indoor line segments to reduce drift, and its current implementation can not completely remove the scale drift by itself. 

The remainder of this paper is organized as follows.  Sec.~\ref{sec:related work} reviews previous related works. Our proposed method is presented in Sec.~\ref{sec:proposed method}, which is then followed by experimental results in Sec.~\ref{sec:experiments}. Sec.~\ref{sec:conclusion} concludes this paper.

\section{RELATED WORK}
\label{sec:related work}
The scale drift in monocular vSLAM system are mainly handled by two approaches: fusion of other sensors and revisit known places.  In \cite{nutzi2011fusion}, two methods, i.e. spline fitting and multi rate Extended Kalman Filter(EKF) are proposed to estimate the scale from an Inertial Measurement Unit (IMU).  A more common approach relies on loop closure with each camera pose represented by 7 DoF rather than 6 DoF in the graph optimization \cite{strasdat2010scale,engel2014lsd}.  Our method leverage the regularity of indoor environment to reduce scale drift without using extra sensors nor relying on revisit known places.  

Some attempts have been made to integrate object recognition results into the SLAM system \cite{tomono2000mobile,fioraio2013joint,botterill2013correcting,galvez2016real}, so the scale information can be obtained from known objects.  These methods require training a detector or classifier ahead, and assume that particular objects can be encountered by the robot, which does not usually occur in a real application scenario.  Our method does not need pre-training, nor access to a model database during running.  The only assumption made in our algorithm is that the line segments in an indoor environment are quantized and form tight clusters, which is usually true in a modern building.

The specialty of indoor environments has been explored in SLAM related research for over a decade \cite{beevers2008inferring,parsley2010towards}.  At `low feature level', the abundant line features in buildings are commonly used \cite{lemaire2007monocular,zhang2009error,zhang2010error,zhang2012vertical,lu2015visual,lu2015robust,zhang2015building,pumarola2017pl} because they are more robust and contain more structure information than key points.  In contrast to using lines for feature matching and structure representation, we observe that not only the appearance of line segments, but the regularity of how they appear can be used to assist SLAM.  Specifically, the frequent existence of line segments of the same length and orientation gives the robot a hint to regulate its map built.

Highly semantic knowledge of the indoor environment are mainly studied for two purpose: using semantic knowledge to assist SLAM, or vice-versa.  A common related research topic is semantic mapping \cite{civera2011towards,stuckler2012semantic,kostavelis2015semantic,kostavelis2016robot}, which targets identifying and recording the meaningful signs in human-inhabited areas.  These approaches take in the map built by the SLAM system together with other inputs, then generate a semantic map that is an enhanced representation of the environment with labels understandable by humans \cite{case2011autonomous,kostavelis2015semantic}.  The highly semantic information, like door numbers and place types (i.e. living room, corridor etc.), are convenient for human-instructed navigation, but it is not easy to use them to benefit SLAM.  It is difficult to integrate information like `scene type' into the formulation of a SLAM system, and practical issues like camera's view point, resolution, stability can hardly guarantee reliable recognition of door number etc.  Our method is based on the assumption that standardized line segments of the same length and orientation appear frequently indoor, so that the line segments observed by the robot form tight clusters.  This knowledge is more abstract and impose more global constraints than local line feature, yet it is not too highly semantic to formulate in the graph optimization.  The performance of proposed method depends on how frequent the line segments within the same cluster appear, whether there is occlusion etc.  One cluster may not be able to regulate the whole trajectory, but as long as its member line segments are detected, our method can help to reduce the scale drift to some extent.

\section{PROPOSED METHOD}
\label{sec:proposed method}
\begin{figure*}
	\centering
	\includegraphics[width=0.75\textwidth]{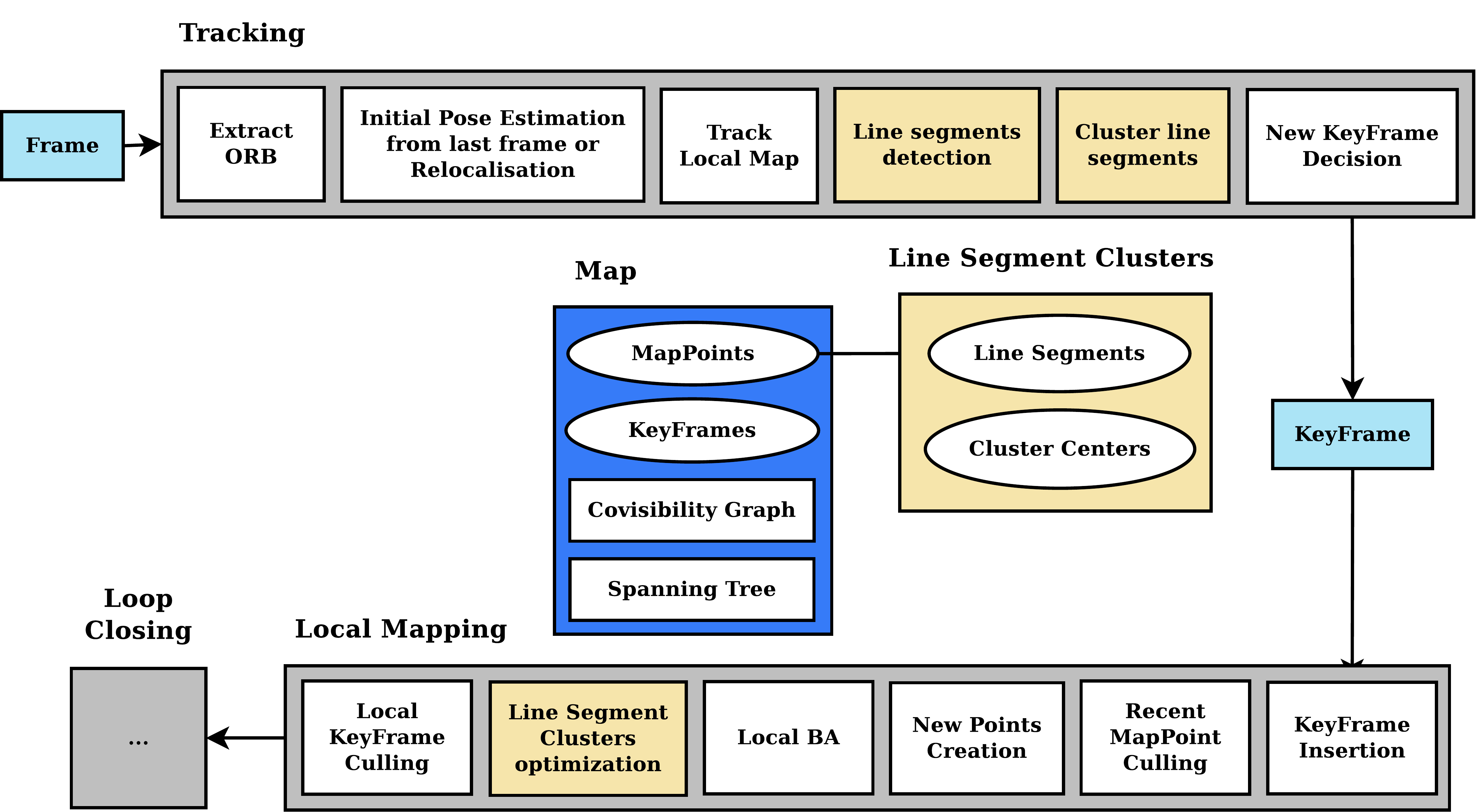}
	\caption{The updated ORB-SLAM2 \cite{mur2015orb,mur2017orb} system with the proposed method integrated.  The`loop closing' thread and place recognition are omitted.  The modules added by our method are shown in light orange.  In the front end, we detect and cluster the line segments in each frame incrementally.  In the back end, we optimize the map imposing the constraint of clustering information. The information maintained by the system is shown in the center.}
	\label{fig:system overview}
\end{figure*}

As mentioned previously, the key observation and the assumption made in our method is that in the indoor environment there exists many quantized line segments that form tight clusters.  By leveraging these line segment clusters the scale drift in monocular SLAM can be reduced.  Our implementation is based on ORB-SLAM2 \cite{mur2015orb,mur2017orb}.  In the tracking thread, our method detects line segments in each frame,  and builds clusters agglomeratively.  In the local mapping thread, after local bundle adjustment, we construct another optimization graph, which contains the map points corresponding to the detected line segments' ends, and imposes the constraint that the segments within one cluster have to be the same.  The updated ORB-SLAM2 \cite{mur2015orb,mur2017orb} system is shown in Figure~\ref{fig:system overview} (better viewed in color), where the modules added by our method are shown in light orange, and the details of the `loop closing' thread and place recognition are omitted.  The original ORB-SLAM2 \cite{mur2015orb,mur2017orb} maintains two things: the place recognition database, and the map which includes map points and key frames.  With our method equipped, we also need to maintain the line segment clusters. (The information stored for a cluster is shown in the left part of Figure~\ref{fig:cluster}.)  

The proposed method consists of three main steps : line segment detection, building clusters, and graph optimization to update the map.  The details of each step are given in the following three subsections.

\subsection{Line Segment Detection}
\begin{figure*}
	\centering
	\includegraphics[width=0.23\textwidth,height=0.13\linewidth]{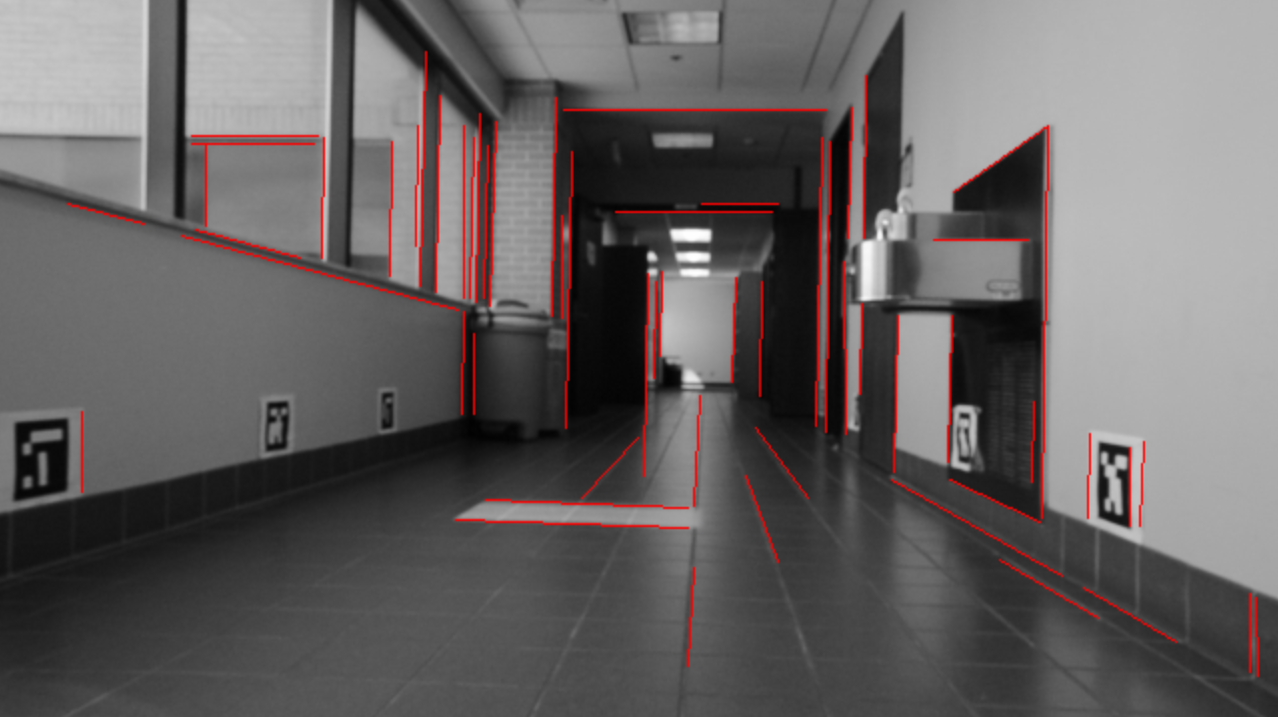}
	\includegraphics[width=0.23\textwidth,height=0.13\linewidth]{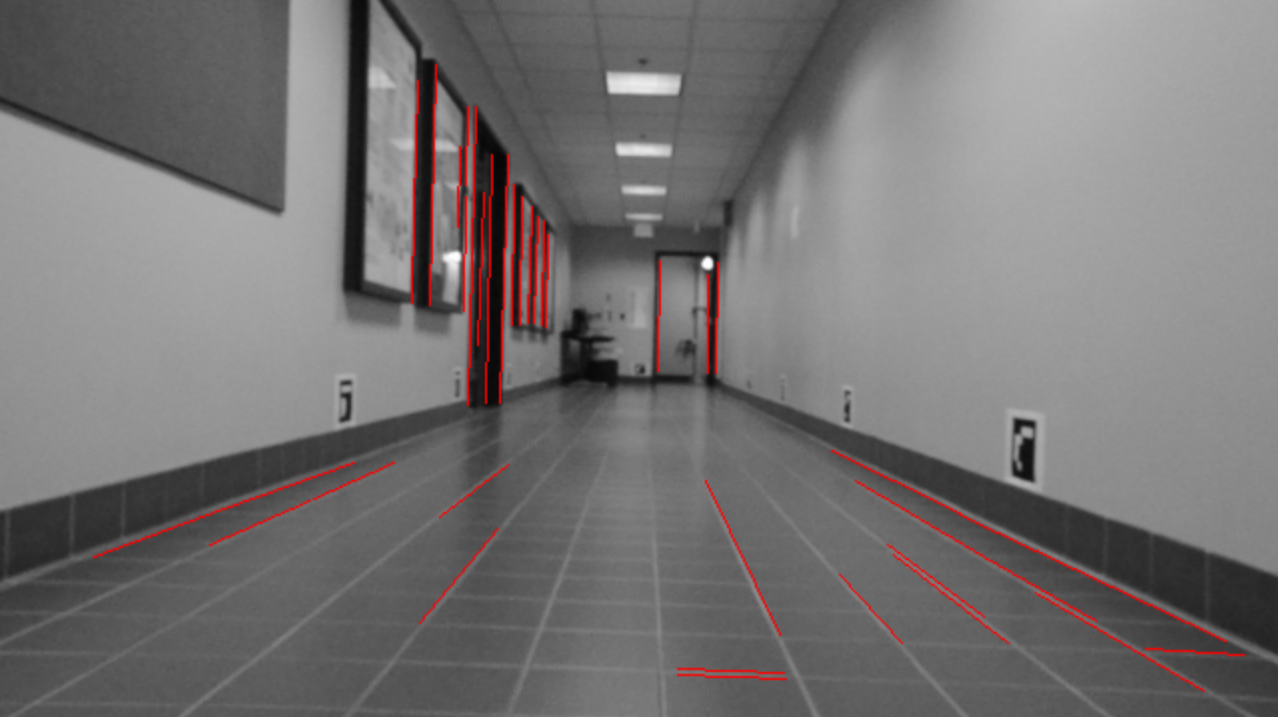}
	\includegraphics[width=0.23\textwidth,height=0.13\linewidth]{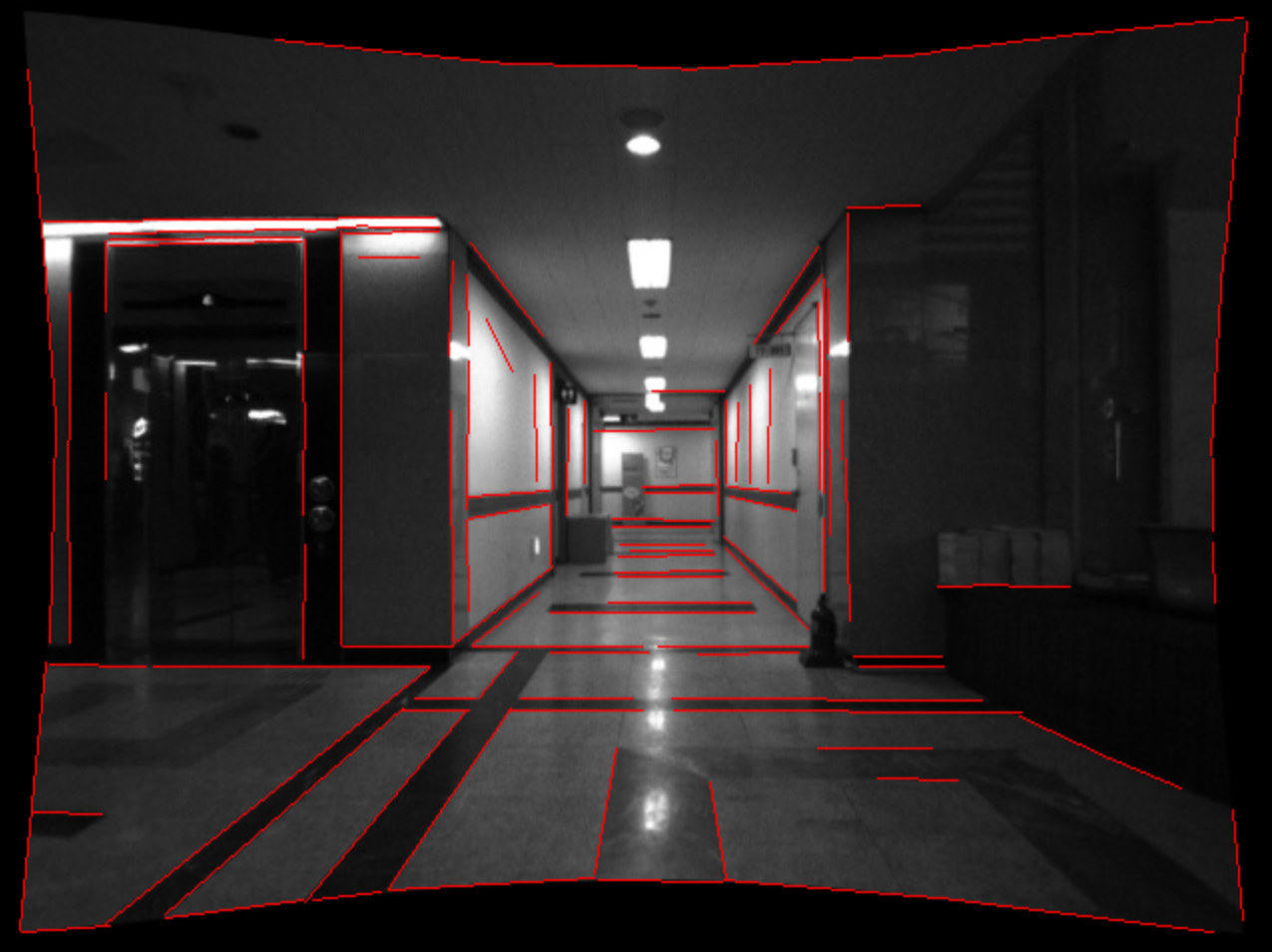}
	\includegraphics[width=0.23\textwidth,height=0.13\linewidth]{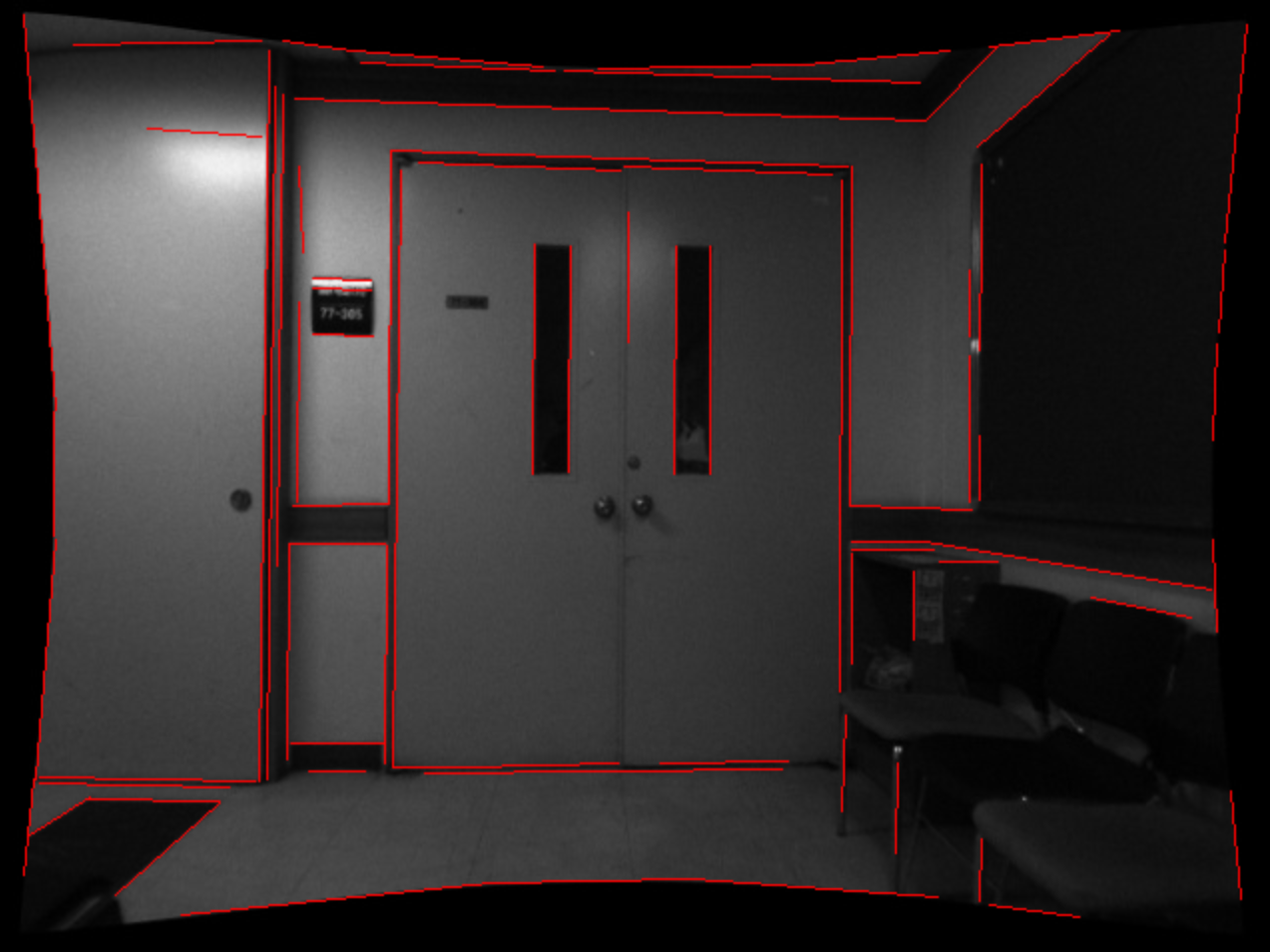}  \\
	\includegraphics[width=0.23\textwidth,height=0.13\linewidth]{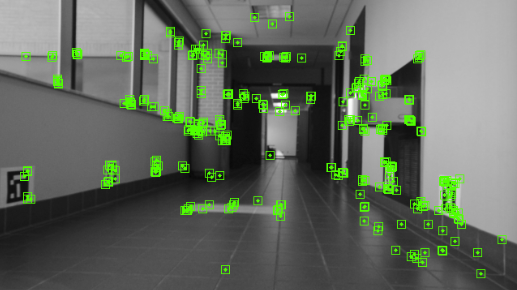}
	\includegraphics[width=0.23\textwidth,height=0.13\linewidth]{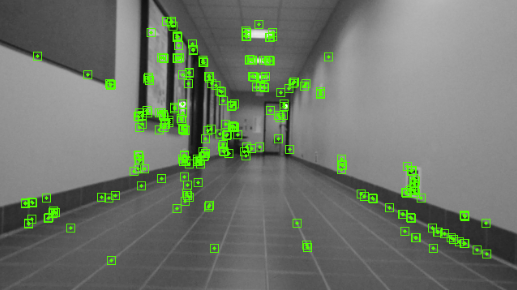}
	\includegraphics[width=0.23\textwidth,height=0.13\linewidth]{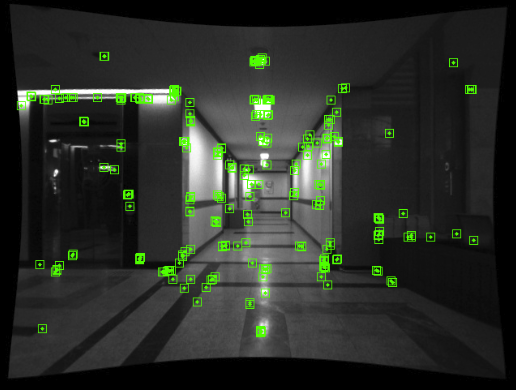}
	\includegraphics[width=0.23\textwidth,height=0.13\linewidth]{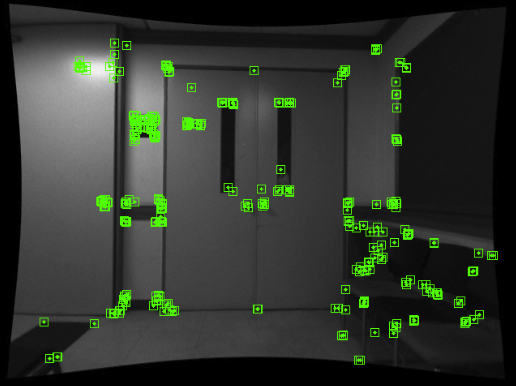}  \\
	\includegraphics[width=0.23\textwidth,height=0.13\linewidth]{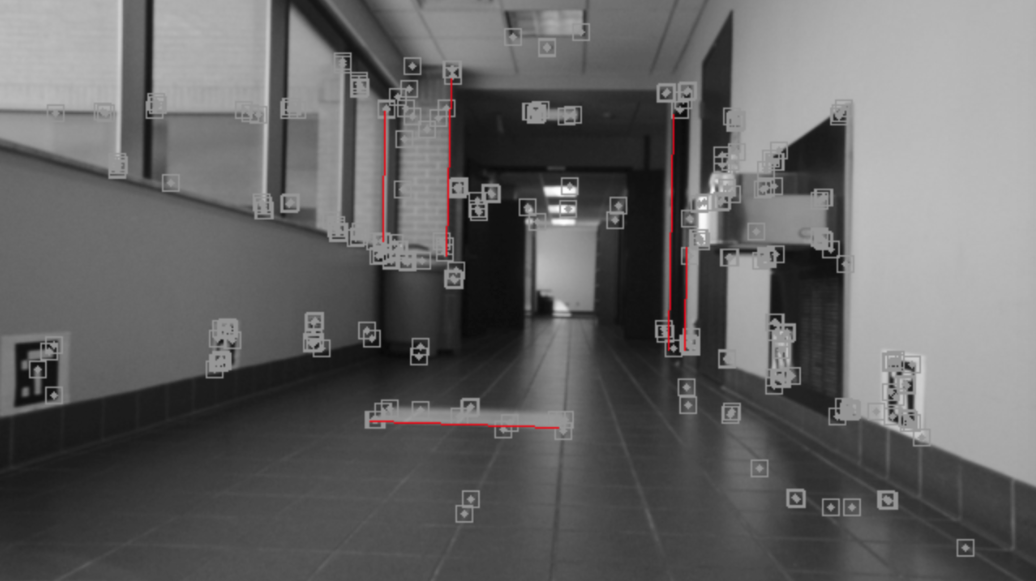}
	\includegraphics[width=0.23\textwidth,height=0.13\linewidth]{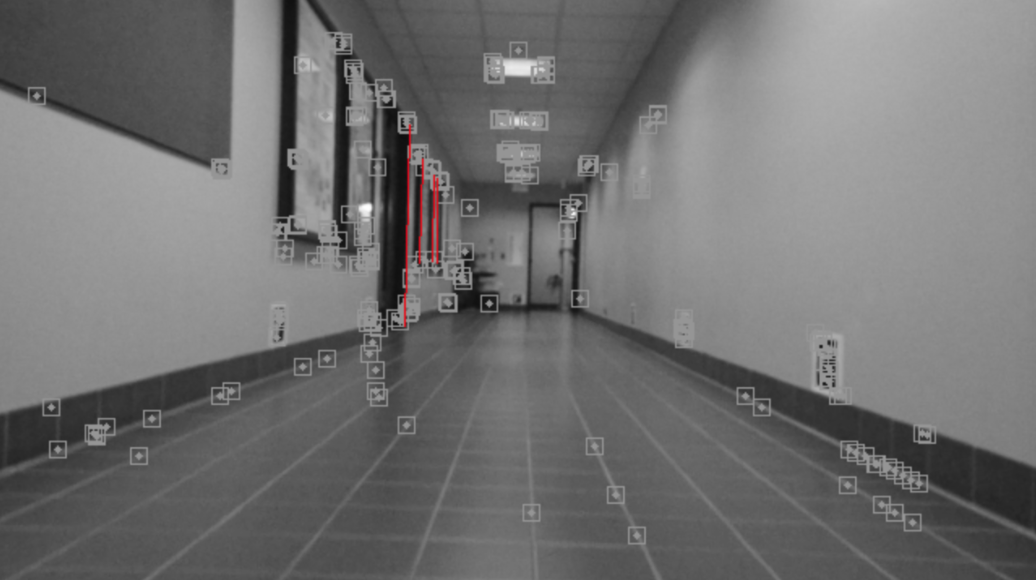}
	\includegraphics[width=0.23\textwidth,height=0.13\linewidth]{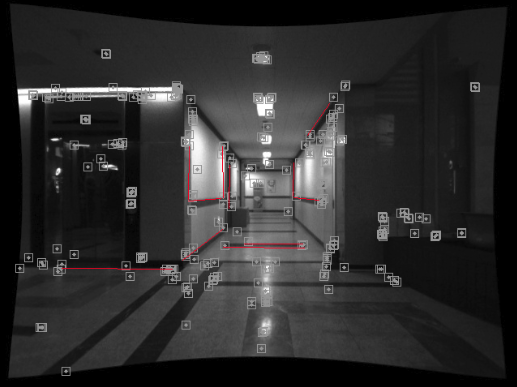}
	\includegraphics[width=0.23\textwidth,height=0.13\linewidth]{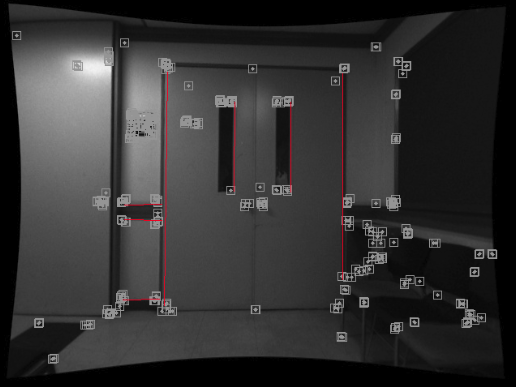}  \\	
	\caption{The first row of images shows the line segment detection with those that are too short and truncated removed.  The second row of images shows the key points detected by ORB-SLAM.  The third row of images shows our final line segment detection results. }
	\label{fig:Line segment detection}
\end{figure*}

We adopt the LSD (Line Segment Detector) \cite{von2010lsd} in our algorithm.  The outputs of the LSD are filtered in the image domain.  The line segments that are either too short (i.e. shorter than 1000 in the image coordinate system) or are truncated by the frame boundary are omitted first.  A problem for the LSD is that it cannot detect the end points of a line segment precisely, as shown in the first row of images in Figure~\ref{fig:Line segment detection}.  By taking advantage of ORB-SLAM2 \cite{mur2015orb,mur2017orb}, we select the line segments whose end points coincide with the detected key points.  The second row in Figure~\ref{fig:Line segment detection} shows the corresponding frames with key points marked, and the last row shows the final line segment detection results in image space, which are then identified and clustered in 3D space (detailed in the next subsection).  Notice that our method does not rely on all the line segments being detected in a frame, nor one line segment being detected in all the frames it appears.  In order for a line segment to contribute in the constraints, it only needs to be detected once in the video sequence it appears. 

\subsection{Building Line Segment Clusters}
Building line segment clusters is conducted in the 3D space that is constructed by ORB-SLAM2 \cite{mur2015orb,mur2017orb} when the map is initialized.  Since we select the line segments whose end points coincide with the detected key points, and ORB-SLAM2 \cite{mur2015orb,mur2017orb} already handles the projection from 2D to 3D for all the key points, given the $i^{th}$ line segment in image space, we obtain its 3D end points $\mathbf{p}_1^i$, $\mathbf{p}_2^i$ for free.  In our method, a line segment is represented by a 3D vector $\mathbf{v}_{seg}^i$, which is the difference between its end points: $\mathbf{v}_{seg}^i = \mathbf{p}_2^i - \mathbf{p}_1^i$.  The center of $j^{th}$ cluster $\mathcal{C}^j$ is the mean vector $\mathbf{v}_c^j$ of its elements:

\begin{equation}
	\mathbf{v}_c^j = \frac{1}{|\mathcal{C}^j|} \sum_{\mathbf{v}_{seg}^i \in \mathcal{C}^j}\mathbf{v}_{seg}^i 
\end{equation}

where $|\mathcal{C}^j|$ is the cardinality or number of elements of cluster $\mathcal{C}^j$.  $\mathbf{v}_c^j$ is incrementally updated when a new element is added.  When we are to decide if a newly detected line segment $\mathbf{v}_{seg}^i$ belong to some existing cluster $\mathcal{C}^j$, the criteria is whether the Euclidean norm of the difference between this vector and the cluster center is within 0.5\% of the length of the cluster center: $||\mathbf{v}_{seg}^i - \mathbf{v}_c^j|| < 0.005 * || \mathbf{v}_c^j|| $.  Since there is no guarantee of the alignment of the directions of $\mathbf{v}_{seg}^i$ and $\mathbf{v}_c^j$, both are tried and the minimum distance is considered.  If $\mathbf{v}_{seg}^i$ is decided to belong to $\mathcal{C}^j$, its end points $\mathbf{p}_1^i$,$\mathbf{p}_2^i$ are stored in $\mathcal{C}^j$, and $\mathbf{v}_c^j$ is updated.  If no existing cluster is for $\mathbf{v}_{seg}^i$, a new cluster is created for it with itself being the cluster center.  The Euclidean distance is directly used to determine the clustering without countermeasure for the drift problem, because the assumption is that the drift happens gradually, while the standardized line segments appear frequently.  Hopefully the elements of the same cluster are identified frequently enough to correct slight drift before a large error accumulates.

\subsection{Graph Optimization of Line Segment Clusters}
\label{subsec:graph opt}
\begin{figure}
	\centering
	\includegraphics[width=0.5\textwidth]{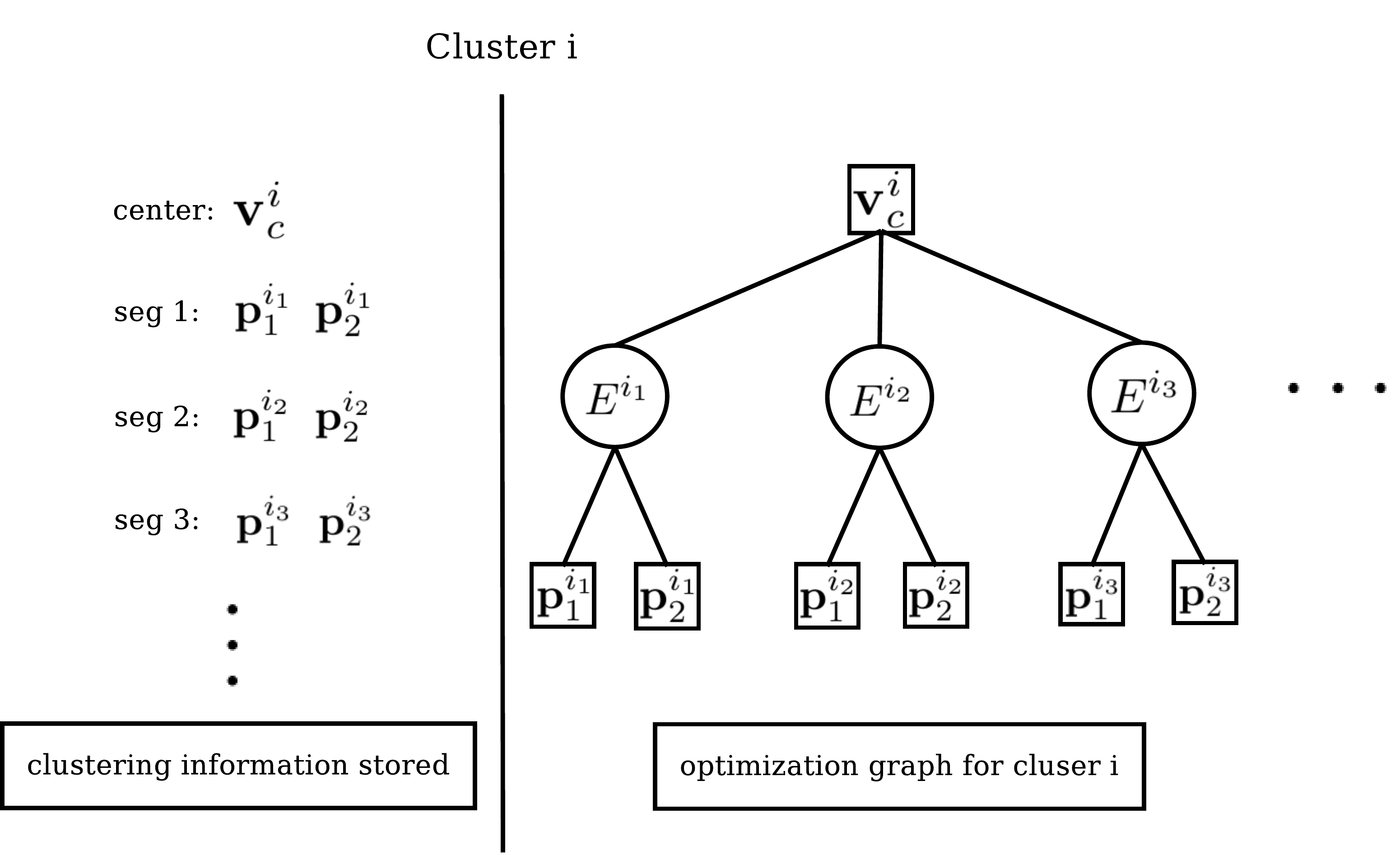}
	\caption{The left part of this figure shows the information stored for cluster i, and the right part shows the graph constructed for this cluster.}
	\label{fig:cluster}
\end{figure}

The clustering information stored is our semi-semantic knowledge, which propagates to the SLAM system by updating the map points involved.  ORB-SLAM2 \cite{mur2017orb} extracts a part of g2o \cite{kummerle2011g} to conduct bundle adjustment.  We replace it with the full version of g2o \cite{kummerle2011g} and define a new type of edge connecting two end points (map points) of a line segment and its cluster center.  The error vector is the difference between the line segment and its cluster center.  Figure~\ref{fig:cluster} shows the information stored for cluster $i$ and the optimization graph constructed for it.  The error vector of edge $E^{i_j}$ is 

\begin{equation}
\label{eqy:e}
	\mathbf{e}^{i_j} = \mathbf{v}_c^i - (\mathbf{p}_2^{i_j} - \mathbf{p}_1^{i_j})
\end{equation}

and the overall objective function is

\begin{equation}
	\underset{\mathbf{p}_1^{i_j}, \mathbf{p}_2^{i_j}}{\mathrm{argmin}} \sum_{i} \sum_{j} (\mathbf{e}^{i_j})^{T} \mathbf{e}^{i_j}.
\end{equation}

We claim that the above objective function helps to reduce the scale drift, impose limited constraint on orientation and have no effect on translational error.  The analysis is as follows.

Suppose $\hat{\mathbf{p}}_1^{i_j}$, $\hat{\mathbf{p}}_2^{i_j}$ are the map points constructed by the SLAM, and relate to their ground truth position $\mathbf{p}_1^{i_j}$, $\mathbf{p}_2^{i_j}$ by

\begin{equation}
\label{equ:p1}
	\hat{\mathbf{p}}_1^{i_j} = s*R*\mathbf{p}_1^{i_j} + T
\end{equation}

\begin{equation}
\label{equ;p2}
	\hat{\mathbf{p}}_2^{i_j} = s*R*\mathbf{p}_2^{i_j} + T.
\end{equation}

The same rotation $R$, translation $t$ and scaler $s$ are used in both (\ref{equ:p1}) and (\ref{equ;p2}) since they are observed in the same frame, i.e. related to one camera pose.  From (\ref{eqy:e}) it can be seen that the line segment clustering constraint applies to the difference between two map points:

\begin{equation}
\begin{split}
\mathbf{v}_c^i & =  \hat{\mathbf{p}}_2^{i_j} - \hat{\mathbf{p}}_1^{i_j} \\
               & =  s * R * (\mathbf{p}_2^{i_j} - \mathbf{p}_1^{i_j}). 
\end{split}	
\end{equation}

The translation $T$ is eliminated.  The rotation $R$ remains but a vector is invariant by rotation with respect to it.  Unfortunately this is a common case in a corridor, e.g. many door frame edges are vertical line segments which are parallel to the axis of ration of the corridor turning.   This explains why our proposed method is particularly helpful in reducing scale drift. 

After the optimization, all the involved map points (i.e. all the $\mathbf{p}$s) are updated according to the clustering constraint.  We conduct our clustering graph optimization after local bundle adjustment, so the map is optimized alternatively to minimize re-projection error and the clusters' variance.  We separate these two graph optimizations for the sake of implementation simplicity, since the map points involved in the clustering graph may be out of the concern of local bundle adjustment, and adding them could ruin the sparse structure of the graph for local bundle adjustment.

\section{EXPERIMENTS}
\label{sec:experiments} 
The effectiveness of our method is shown by comparing of the trajectories built by a monocular SLAM system with and without the assist of the proposed line-cluster optimization.  We pick the test sequences that are taken in a typical corridor without loop closure.

\subsection{Implementation}
Our implementation is based on ORB-SLAM2 \cite{mur2017orb}, which also serves as our baseline.  With the proposed method integrated, the augmented SLAM system still runs in real time.  The results of the proposed method are presented with 2 different configurations named \textit{Seg} and \textit{SegGlobal}.  In \textit{Seg}, local bundle adjustment and clustering optimization are conducted alternatively, while in \textit{SegGlobal}, additional global bundle adjustment is conducted after each clustering optimization, helping to propagate the semi-semantic knowledge to the whole map built. 

We adopt the widely used absolute trajectory error (ATE) \cite{sturm2012benchmark} and relative pose error (RPE) \cite{sturm2012benchmark} as our numeric measurement for location only.  The obtained camera pose trajectories are aligned to the ground truth by finding the transformation that minimizes their sum of squared error.

\subsection{Datasets}
Finding the proper test sequences turn out to be challenging, and there are two main problems: 1) missing ground truth and 2) the quality of the video content.  For the trajectories of very small scale, i.e. within one room, the ground truth can be collected by optical tracking system or markers deployed in the room; for the outdoor environments, GPS is available; while the corridors are neither easy to deploy a system nor have GPS, but that is where vSLAM is most needed.  As for the content, we want to extract a long sequence without loop closure, and the monocular mode ORB-SLAM2 \cite{mur2017orb} will not lose tracking during it.  Some dataset contains sharp turnings in front of blank walls, causing the SLAM system losing tracking frequently.  We show our experimental results on two datasets: HRBB4 \cite{lu2014high} and IT3F \cite{zhang2015building}.  

\begin{figure}
	\centering
	\includegraphics[width=0.15\textwidth]{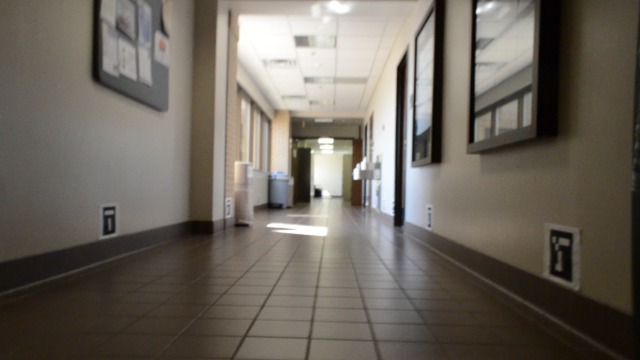}
	\includegraphics[width=0.15\textwidth]{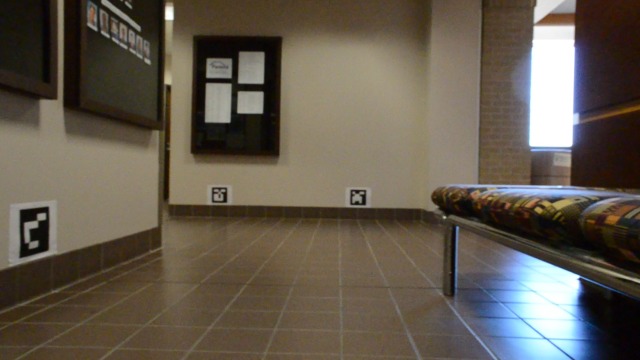}
	\includegraphics[width=0.15\textwidth]{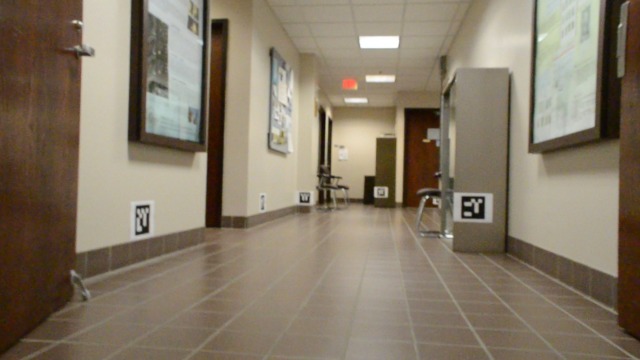}
	\caption{The sample images from HRBB4 dataset \cite{lu2014high}.}
	\label{fig:HRBB4 samples}
\end{figure}

HRBB4 dataset \cite{lu2014high} contains an image sequence of 12,000 frames of 640 x 360 pixels captured from the $4^{th}$ floor of the H.R. Bright Building, Texas A\&M University using a Nikon 5100 camera.  Some sample images are shown in Figure~\ref{fig:HRBB4 samples}. The ground truth camera positions (no orientation) of a subset of frames are offered, and we interpolate both ground truth and the SLAM results by spline interpolation \cite{de1978practical}.

\begin{figure}
	\centering
	\includegraphics[width=0.15\textwidth]{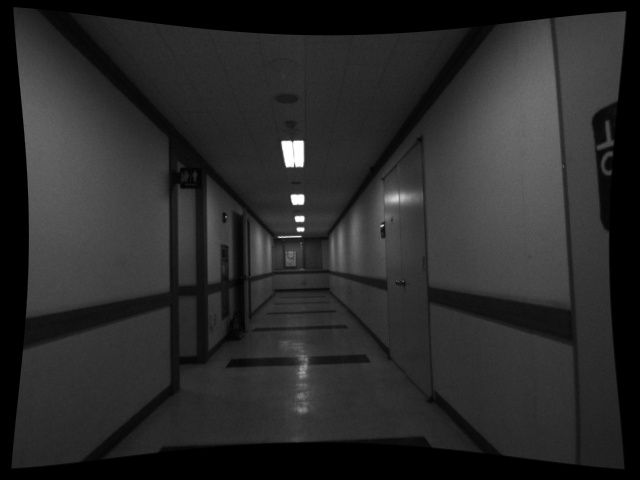}
	\includegraphics[width=0.15\textwidth]{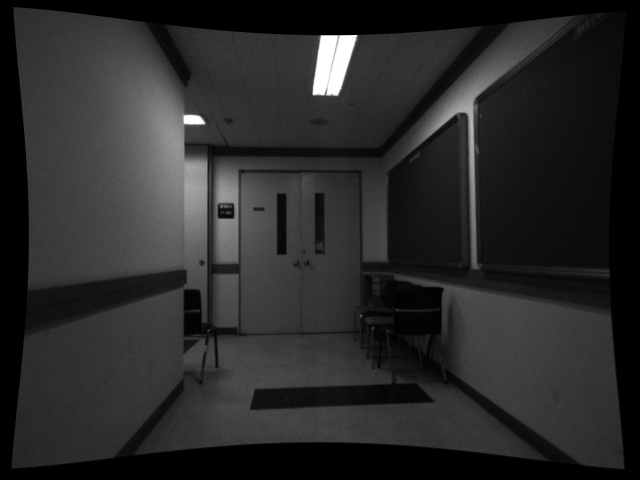}
	\includegraphics[width=0.15\textwidth]{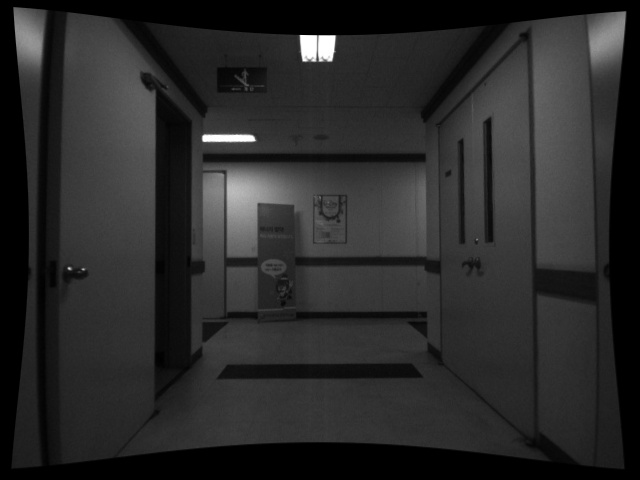}
	\caption{The sample images from IT3F dataset \cite{zhang2015building}.}
	\label{fig:IT3F samples}
\end{figure}

IT3F dataset \cite{zhang2015building} is collected from the $3^{rd}$ floor of the IT building at Hanyang University, with the dimensions $24 \times 11.5$~m.  A calibrated Bumblebee BB2-08S2C-38 is used as the vision sensor, and all the images are undistorted and rectified.  Some sample images are shown in Figure~\ref{fig:IT3F samples}.  This dataset is proposed for stereo SLAM, and it contains multiple loop closures but no ground truth is offered.  The stereo results from ORB-SLAM2 \cite{mur2015orb,mur2017orb} are used as ground truth in our experiments, and we conduct monocular SLAM using an extracted sequence captured by the left camera.

\subsection{Results}

\subsubsection{Results on HRBB4 Dataset}
\begin{figure}
	\centering
	\includegraphics[width=0.45\textwidth]{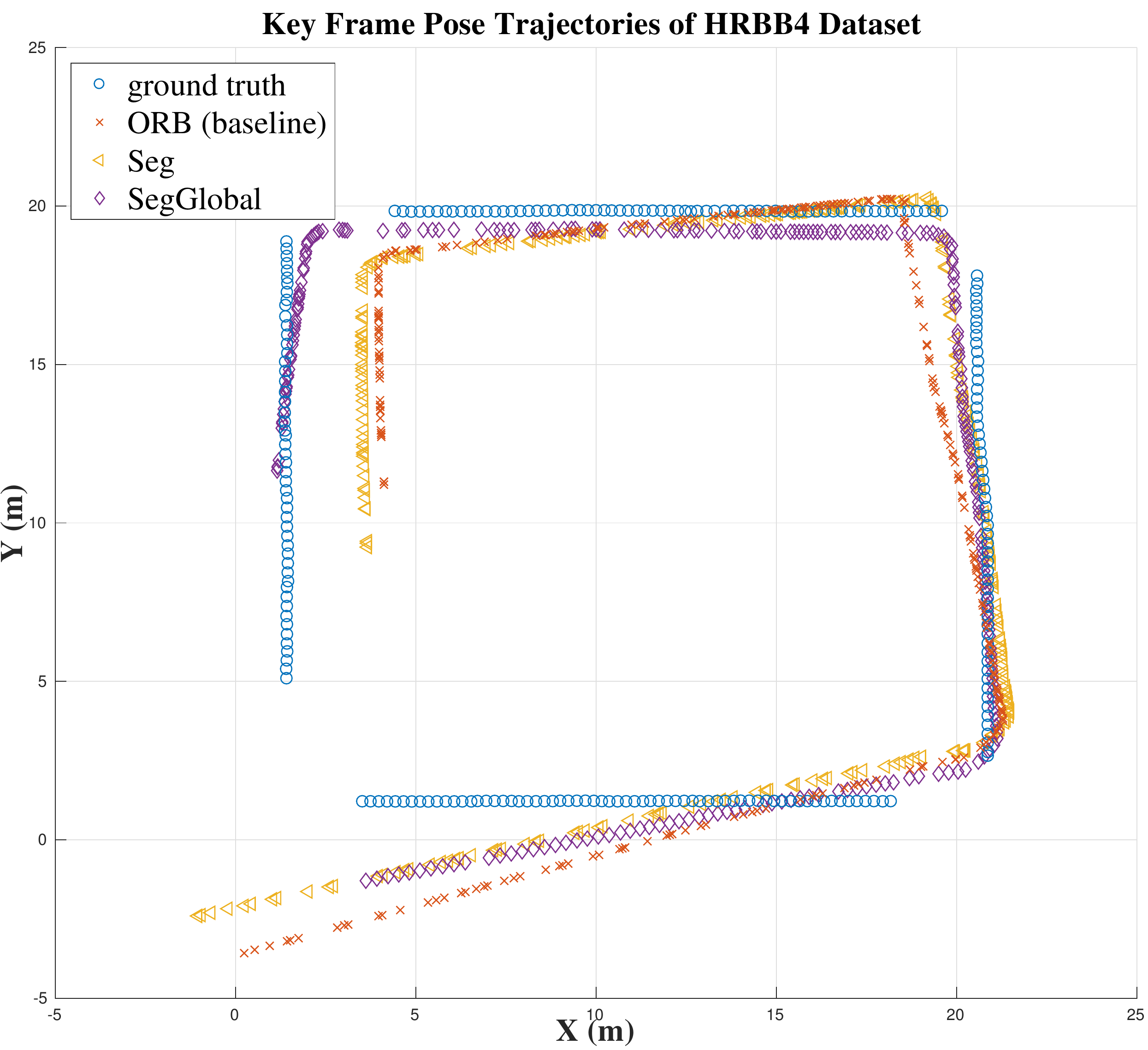}
	\caption{The aligned trajectories of HRBB4 dataset \cite{lu2014high}.}
	\label{fig:HRBB4}
\end{figure}

We use the whole sequence of the HRBB4 dataset \cite{lu2014high} which dose not contain loop closure and the monocular ORB-SLAM2 \cite{mur2017orb} can run through the whole sequence without losing tracking.  The aligned trajectories are shown in Figure~\ref{fig:HRBB4}, and the corresponding ATE and RPE results are shown in Table~\ref{tab:HRBB4}.   It can be seen that our proposed method with configuration \textit{SegGlobal} achieves the best performance, far exceeding our baseline, and with configuration \textit{Seg}, our method still improves the baseline.

\begin{table}[h]
	\begin{center}
		\begin{tabular}{|l|c c c|}
			\hline
		                  &  ORB-SLAM2  & Seg       & SegGlobal \\
			\hline\hline
			ATE           &  $1.9879$   & $1.8248$  & $\mathbf{1.0062}$  \\
			PRE           &  $3.8383$   & $3.5086$  & $\mathbf{1.9801}$ \\
			\hline
		\end{tabular}
	\end{center}
	\caption{The ATE and PRE of HRBB4 dataset \cite{lu2014high}.  The best results achieved are highlighted in boldface.}
	\label{tab:HRBB4}
\end{table}

\subsubsection{Results on IT3F Dataset}
\begin{figure}
	\centering
	\includegraphics[width=0.45\textwidth]{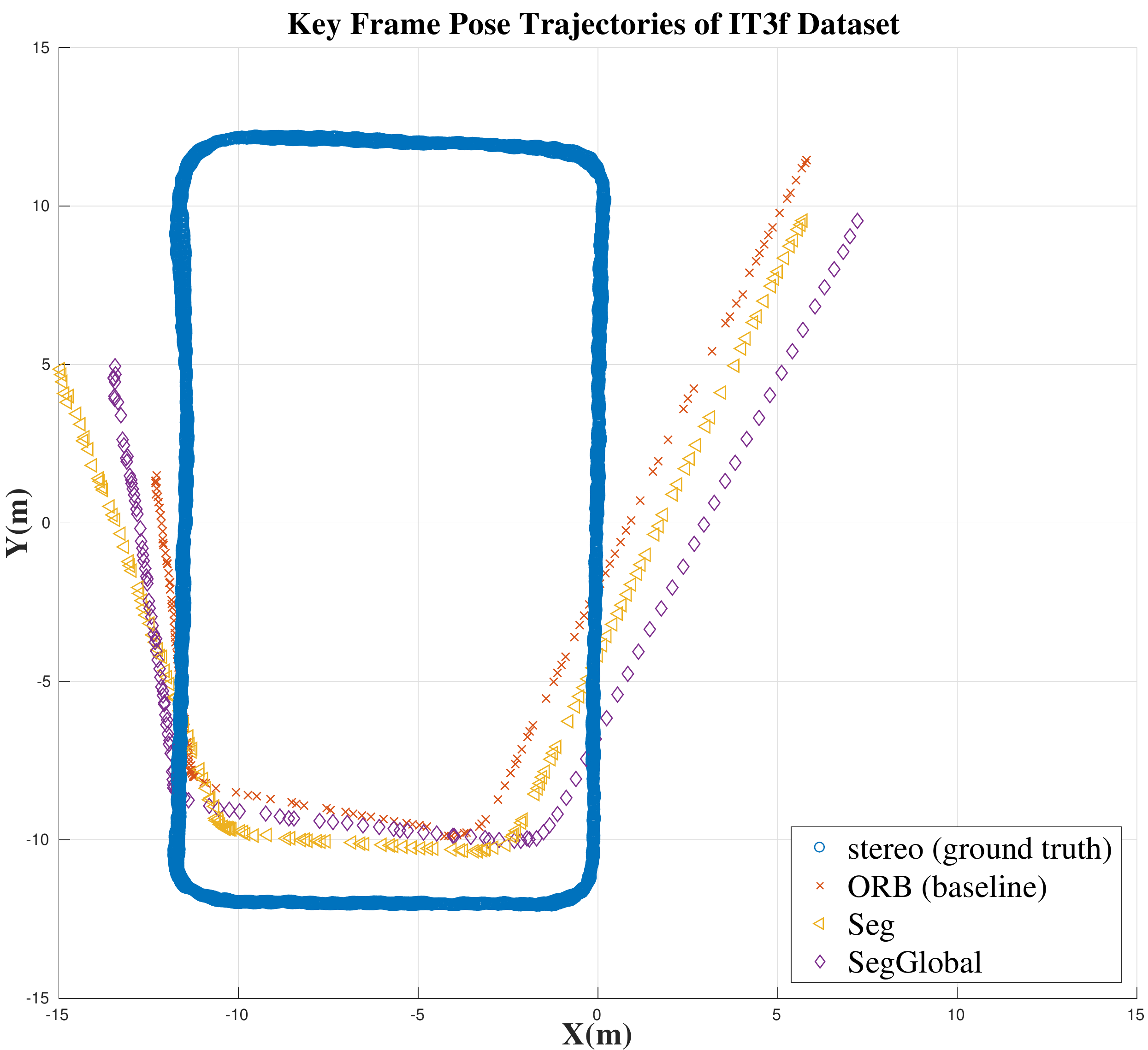}
	\caption{The aligned trajectories of IT3F dataset \cite{zhang2015building}.}
	\label{fig:IT3F}
\end{figure}

The aligned trajectories are shown in Figure~\ref{fig:IT3F}, and the corresponding ATE and RPE results are shown in Table~\ref{tab:IT3F}.  We found that monocular SLAM is incapable of keeping tracking of a complete circle (but stereo SLAM can), which is why there is a missing part in all the results. (Tracking is lost after turning in front of a blank wall.)  We extract the longest sequence (about 4600 frames) that can be followed by monocular ORB-SLAM2 \cite{mur2017orb} to show our results. It can be seen from Figure~\ref{fig:IT3F} that with or without our method equipped, the orientation errors accumulate with each turning alike, but Table~\ref{tab:IT3F} shows that the results with our method are closer to the ground truth.  According to the analysis in subSec.~\ref{subsec:graph opt}, this improvement comes from the reduced scale drift.

\begin{table}[h]
	\begin{center}
		\begin{tabular}{|l|c c c|}
			\hline
		                  &  ORB-SLAM2  & Seg       & SegGlobal \\
			\hline\hline
			ATE           &  $3.5841$   & $\mathbf{2.5380}$  & $2.9918$  \\
			PRE           &  $6.0966$   & $\mathbf{4.4054}$  & $5.1996$ \\
			\hline
		\end{tabular}
	\end{center}
	\caption{The ATE and PRE of IT3F dataset \cite{zhang2015building}.  The best results achieved are highlighted in boldface.}
	\label{tab:IT3F}
\end{table}

\subsubsection{Computational Time}
We run the monocular ORB-SLAM2 \cite{mur2017orb} with the proposed method integrated on a desktop with Intel® Core™ i5-4570 CPU @ 3.20GHz x 4 processors.  For input image of size $480 \times 640$, our method costs about 0.031 seconds per frame in the front end, i.e. the two orange modules in tracking thread in Figure~\ref{fig:system overview}, and about 0.012 seconds per round optimization (10 iterations) in the back end, i.e. the orange module in local mapping thread in Figure~\ref{fig:system overview}.   For your information, the local bundle adjustment of ORB-SLAM2 \cite{mur2017orb} cost about 0.21 seconds per round (15 iterations), while the time cost by global bundle adjustment increases as the map grows.

\section{CONCLUSIONS}
\label{sec:conclusion}
In this paper we proposed a novel method to reduce the scale drift for indoor monocular SLAM.  Our method assists the SLAM system by leveraging the regularity of the abundant standardized line segments in the indoor environment.  We observed that the indoor line segments are quantized and form tight clusters, which can be used to regulate the map built by SLAM.  Our method makes a very weak assumption and does not need pre-training nor to access model database while running, and thus is very generalizable.  We implemented the proposed method in the popular ORB-SLAM2 \cite{mur2017orb} and take advantage of its results to save computation.  In the front end we detect the line segments in each frame and incrementally cluster them in the 3D space.  In the back end, we optimize the map imposing the constraint by our newly defined edges.  The performance of our method depends on how frequent the line segments within the same cluster appear, whether there is occlusion etc.  Experimental results showed that our proposed method successfully reduces the scale drift for indoor monocular SLAM.  


%%%%%%%%%%%%%%%%%%%%%%%%%%%%%%%%%%%%%%%%%%%%%%%%%%%%%%%%%%%%%%%%%%%%%%%%%%%%%%%%

%\section*{ACKNOWLEDGMENTS}

\end{document}